# Brain Stroke Detection and Classification Using CT Imaging with Transformer Models and Explainable AI

Shomukh Qari[1], Maha A. Thafar[1*]

[1] College of Computers and Information Technology, Computer Science Department,
Taif University, Taif, Saudi Arabia.

[*]Corresponding author: (email: m.thafar@tu.edu.sa)

## Abstract

Stroke is one of the leading causes of death globally, making early and accurate diagnosis essential for improving patient outcomes, particularly in emergency settings where timely intervention is critical. CT scans are the key imaging modality because of their speed, accessibility, and cost-effectiveness.

This study proposed an artificial intelligence framework for multiclass stroke classification (ischemic, hemorrhagic, and no stroke) using CT scan images from a dataset provided by the Republic of Turkey's Ministry of Health. The proposed method adopted MaxViT, a state-of-the-art Vision Transformer, as the primary deep learning model for image-based stroke classification, with additional transformer variants (vision transformer, transformer-in-transformer, and ConvNext) used for comparison. To enhance model generalization and address class imbalance, we applied data augmentation techniques, including synthetic image generation. The MaxViT model trained with augmentation achieved the best performance, reaching an accuracy and F1-score of 98.00%, outperforming all other evaluated models and the baseline methods.

The primary goal of this study was to distinguish between stroke types with high accuracy while addressing crucial issues of transparency and trust in artificial intelligence models. To achieve this, Explainable Artificial Intelligence (XAI) was integrated into the framework, particularly Grad-CAM++. It provides visual explanations of the model's decisions by highlighting relevant stroke regions in the CT scans and establishing an accurate, interpretable, and clinically applicable solution for early stroke detection. This research contributed to the development of a trustworthy AI-assisted diagnostic tool for stroke, facilitating its integration into clinical practice and enhancing access to timely and optimal stroke diagnosis in emergency departments, thereby saving more lives.

**Keywords:**
*Stroke Multiclass classification,* ischemic stroke, hemorrhagic stroke, *pretrained models, Transfer learning, vision Transformer, XAI.*

## 1 Introduction

Stroke is one of the leading causes of death and disability worldwide, according to the 2021 World Health Organization (WHO) report. It ranks third globally as a cause of death, accounting for about 10% of annual deaths, which is approximately 6.6 million people each year. There are various causes of stroke, including atherosclerosis, high blood pressure, and lifestyle factors such as lack of exercise, smoking, and alcohol consumption. Strokes are generally divided into two types: ischemic and hemorrhagic. Ischemic stroke is the most common type of stroke, which represents about 87% of all stroke cases ("Global Health Estimates: Life Expectancy and Leading Causes of Death and Disability," n.d.). It occurs due to a lack of blood flow to the brain caused by a blockage in the major blood vessels that a blood clot may block. This blockage leads to a lack of oxygen and nutrients in the brain, and as a consequence, the brain cells and tissues begin to die. The second type is Hemorrhagic stroke, which accounts for only 13% of all strokes. It happens when blood vessels that supply oxygen to the brain cells and tissues rupture and bleed, causing an increased pressure on the surrounding tissue inside the skull, which can further lead to brain damage.

Early stroke detection is the first essential step in stroke management, and it is crucial in improving survival rates, which exceed 90% (Denslow 2022). Several imaging techniques have been utilized for stroke detection and diagnosis. Computed tomography (CT) and magnetic resonance imaging (MRI) are the two dominant imaging modalities for stroke detection and classification, each presenting unique advantages aligned with specific clinical requirements. The time factor is sensitive in stroke detection. Efficient, accurate, and fast stroke identification and determination of its type are critical to improve the patient's outcome and deciding the treatment plan. Misclassification of the stroke type causes incorrect treatment since ischemic stroke requires treatments to restore blood flow for vessel blockage, while hemorrhagic stroke requires actions to control and stop bleeding (Gomes and Wachsman 2013). Therefore, misclassification can lead to harmful consequences and cause death.

Early detection of stroke increases the survival rate, and the faster and more accurate the discovery, the higher the chance of survival. Since stroke usually occurs quickly, it may be difficult to notice the patient's symptoms, and when the symptoms appear clearly, the time factor becomes very sensitive. Therefore, physicians must intervene fast and get the examination results immediately. For this early detection of stroke, several brain imaging modalities can be used, such as MRI and CT scans (Sirsat, Fermé, and Câmara 2020). Despite using both medical imaging techniques in early stroke detection, CT scans remain the standard imaging in emergency settings because of their cost-effectiveness, accessibility, and availability in most hospitals and healthcare providers, allowing for immediate assessment. Hemorrhagic stroke can often be detected smoothly through CT scans, but CT scans have limitations in detecting ischemic changes in the early phases. Therefore, there is an urgent need for a precise, automated, and effective diagnosis system that supports physicians in analysis, investigation, and stroke identification.

This research addressed the challenge of stroke detection in two aspects. The first aspect is developing an automated AI-based system that leverages CT scan data and advanced AI techniques integrated together in different stages of a complete pipeline to detect the presence of a stroke and distinguish between ischemic and hemorrhagic types in real time. The second aspect is facing some limitations in implementing a smart automatic system, such as the limited size of the training data and the need for results and diagnosis interpretability. Most medical professionals need clarification of the results to make them more logical for them and to reduce their doubts about their accuracy, and this can be done through the use of Explainable AI (XAI).

This research contribution folds into four main points:
1. Provides an automatic system for accurate and fast stroke detection using CT-scan imaging.
2. Automatically identify the type of stroke by implementing a multiclass classification method to distinguish between strokes: hemorrhagic, ischemic, or normal, enhancing diagnostic precision.
3. Leverages advanced AI techniques by utilizing the Hybrid-based and transformer-based pretrained models for stroke detection.
4. Integrates the XAI to improve the transparency and interpretability of model predictions, aiding clinicians in understanding the model's decision-making process.

The remainder of this research is structured as follows: Section 2 provides background by explaining the medical-related and AI-related terms and concepts. Section 3 summarizes the research relevant literature and outlines the ML, DL, and transfer learning (TL) methods developed for stroke detection. Section 4 describes the datasets utilized in this study and introduces this study methodologies. Section 5 explains the experimental design, covering the evaluation metrics, training and testing process. Section 6 discusses the results and key findings. Finally, section 7 concludes the study and highlights future work.

## 2 Background and Concept

Understanding the key features of stroke and differentiating between each type is crucial for designing an AI model capable of detecting stroke and distinguishing its type based on the unique and hidden pattern recognized by advanced DL techniques. Therefore, before delving into the literature, we first introduce some necessary concepts from both medical and AI perspectives, which will facilitate a better understanding of the literature and the methodology discussed in the following chapter.

## 2.1 Medical Imaging-related Concepts for Stroke Detection

### *2.1.1 Stroke Overview*

Brain stroke is one of the diseases whose early detection is of high importance, and stroke can be classified generally into two types.

**Ischemic stroke** represents the largest part (87%) of the total cases and occurs due to ischemia of the brain (Powers 2020). A stroke of this type is caused by intracranial thrombosis or extracranial embolism, which travels through the bloodstream from the heart, often. An ischemic stroke increases the size of the infarction, and an infarction is an area where cells have been damaged, resulting in dead tissue. Infarction is the main feature of ischemic stroke. The infarction can be observed using a CT scan, and the size of the infarction varies based on the percentage of dead tissue in the affected area of the brain, where the size of the infarction is directly proportional to the increase in the persistence of ischemia.

**Hemorrhagic stroke** occurs due to bleeding in the brain resulting from the rupture of blood vessels. Stroke of this type is classified as intracerebral hemorrhage (ICH) and subarachnoid hemorrhage (SAH) (Montaño, Hanley, and Hemphill 2021). Hemorrhagic stroke is often caused by high blood pressure. A hemorrhagic stroke can be determined by observing bleeding in the brain that can lead to hematoma, and hematoma is the pool of blood in the area where the bleeding occurs. Bleeding is the primary feature of hemorrhagic stroke and can also be observed with CT scans and falsity appears in CT as a high-intensity area.

### *2.1.2 Key Imaging Features Relevant to Stroke Diagnosis*

Accurate detection and classification of stroke types rely on identifying specific patterns in brain imaging (Shafaat and Sotoudeh 2023). Two major indicators are infarction and bleeding, which help distinguish between ischemic and hemorrhagic strokes. In the case of an ischemic stroke, a region of the brain may appear darker on a CT scan, indicating a low-density (hypodense) area. This suggests tissue damage caused by a blocked blood vessel, commonly referred to as an infarction (Shafaat and Sotoudeh 2023). On the other hand, hemorrhagic strokes are characterized by bleeding within or around the brain. In CT scans, this bleeding typically appears as a brighter, high-density (hyperdense) area. For example, when bleeding occurs in the subarachnoid space, it may indicate a subarachnoid hemorrhage (SAH). If the bleeding is located within the brain tissue itself, it is likely an intracerebral hemorrhage (ICH). The location and intensity of these bright regions help in identifying the specific type of hemorrhagic stroke.

These distinctive imaging features and other important features related to the brain structure and stroke location summarized in Table 1 (Karthik et al. 2020) are vital for DL pretrained models to detect and distinguish between ischemic and hemorrhagic strokes accurately, enabling timely intervention and treatment.

**Table 1:** Key Differences between hemorrhagic and ischemic Stroke

| Feature | Hemorrhagic Stroke | Ischemic Stroke |
| --- | --- | --- |
| **CT Appearance** | Hyperdense (bright) regions | Hypodense (dark) regions |
| **Lesion Markers** | - Blood accumulation<br>- Hematoma | - Infarcted tissue<br>- Reduced perfusion |
| **Midline Shift** | Common due to mass effect | Less common, unless large infarct |
| **Possible Locations** | Basal ganglia, cerebellum, brainstem | Middle cerebral artery territory |
| **Detection** | - Rapid onset<br>- Easily detected on CT | - Often detected later.<br>- Subtle signs |

## 3 Related Work

Over the past years, stroke accounted for a large proportion of the number of deaths around the world. Traditionally, its diagnosis has relied on a combination of clinical assessments, neuroimaging techniques such as CT and MRI, as well as blood tests. However, there has been a growing shift towards the integration of artificial intelligence (AI) in the medical and healthcare sectors (Thafar et al. 2023; Albaradei et al. 2023, 2021, 2022; Alamro et al. 2023). One of these applications is the early detection of stroke, where CT and MRI imaging have been utilized along with DL and ML techniques. Given the limitations in current stroke treatment, early detection is essential to improving patient outcomes. AI-driven methods can enhance stroke identification by reducing diagnostic delays and enabling timely intervention.

This section reviews recent research (2020–present) on early stroke detection, categorizing the studies into three methodological approaches: Deep Learning, Machine Learning, and Transfer Learning. These categories highlight key advancements and emerging trends in the field.

### 3.1 Stroke Detection Techniques Using Machine Learning

Several studies utilized machine learning for stroke detection. The first study utilized ML techniques to detect stroke is done by Sirsat and coauthors (Sirsat, Fermé, and Câmara 2020) This study reviewed 39 published research with the aim of classifying the latest ML techniques for stroke detection into four categories based on their functions or similarities, which are:stroke prevention, stroke diagnosis, stroke treatment, and stroke prognostication/outcome prediction. The related category to our study is the second one, where CT scans were used as a database repeatedly in studies. According to the findings, the support vector machine (SVM) classifier achieved the best results in ten studies, while both SVM and Random Forest (RF) proved their effectiveness across all studies. Another notable study (Peixoto and Rebouças Filho 2018) presented a new approach to the classification of stroke subtypes, specifically hemorrhagic and ischaemic strokes, using SVM, Multi-Layer Perceptron (MLP), Minimal Learning Machine, Linear Discriminant Analysis (LDA), and Structural Co- Occurrence Matrix (SCM). According to this study, SCM is the best to extract the most discriminant structural information concerning stroke subtype without requiring parameters tuning. The authors used a dataset of 300 CT scans images.The SCM in the frequency domain presented high accuracy i.e. 98% in stroke classification. Despite the promising results reported in several studies that utilized ML for stroke identification, the ML approaches still have multiple limitations and challenges. For example, ML heavily depends on manual feature engineering (extraction and selection), which often requires domain expertise and it is a time-consuming process. Moreover, Traditional ML models often struggle to scale with large and complex datasets, especially when investigating high-dimensional CT images. These limitations justify the transition to DL models.

## 3.2 Deep Learning Approaches for Stroke Detection

Several studies have discussed early stroke detection using the DL methodology. One of these studies (Wang et al. 2021) discussed the prediction of hemorrhagic stroke using CT scans obtained from the 2019-RSNA Brain CT Hemorrhage Challenge. The data was gathered from three institutions: Stanford University, Universidade Federal de São Paulo Brazil, and Thomas Jefferson University Hospital Philadelphia, and subsequently re-annotated by the American Society of Neuroradiology with the assistance of over 60 neuroradiologists (Flanders et al. 2020). The authors designed a DL approach that mimics the interpretation process of radiologists by combining a 2D CNN model and two sequence models to achieve accurate detection and subtype classification of acute intracranial hemorrhage (ICH). The model achieved high prediction performances in terms of AUC for the detection of different types of bleeding as it achieved the following results: (ICH): 0.988, (EDH): 0.984, (IPH): 0.992, (IVH): 0.996, (SAH): 0.985 and (SDH): 0.983.

Another study (Gautam and Raman 2021) utilized CNN to classify CT scans into hemorrhagic, ischemic stroke and normal using a new 13-layer CNN architecture called P-CNN. Their model was found to be better than the well-known traditional CNN architectures such as AlexNet and ResNet50. P_CNN consists of input layer,convolution layer followed by a rectified linear unit (ReLU) activation function, Max pooling, and Dropout layers to prevent the overfitting., The last layer is fully-connected and uses softmax to normalize the output in range [0 1] for classification task. The model was tested on two different CT scan datasets. The classification accuracy obtained by their method in the first experiment is 98.33% and in the second experiment is 98.77%. Similarly the study in (Qiu et al. 2020) developed an automated approach to detect and quantitate infarction by using non–contrast-enhanced CT scans in patients with acute ischemic stroke. A ML approach to segmentation of infarction areas in non-optimized CT images in acute ischemic stroke patients showed good compatibility with stroke size on water-based MRI images.

While DL-based methods have enhanced stroke detection and classification by automating feature extraction with high prediction accuracy, they still face some challenges, such as the need for a massive dataset to train the model from scratch, which is very time-consuming and has high computational costs. Therefore, TL has emerged to eliminate some of these issues.

## 3.3 Transfer Learning with Pretrained Models (CNN/ Transformer)

Transfer learning (TL) has emerged as a practical solution to overcome several limitations associated with DL approaches, particularly by leveraging pretrained models tailored for vision tasks in the medical imaging domain. Notably, Vision Transformer (ViT)-based architectures have been increasingly adopted in disease diagnosis from medical images, demonstrating promising results across various applications (Manzari et al. 2023; Alghoraibi et al. 2025; Alharthi and Alzahrani 2023; Alzahrani 2025). Recent studies that utilized TL with pretrained models to advance stroke detection and classification tasks can be categorized to CNN-based and Transformer-based architectures. One of these studies (Kaya and Önal 2023) developed a novel CNN architecture for detecting and classifying brain stroke into two classes ((hemorrhagic and ischemic) and three classes (hemorrhagic, ischemic, and normal) using non-contrast brain CT images. CNN and U-Net were employed to classify strokes and normals, and then image segmentation was utilized to detect the stroke type. The model's architecture consists of 19 layers, and the model achieved a 95.06% for the classification model.

One notable study (Çınar, Kaya, and Kaya 2023) demonstrates the effectiveness of TL by utilizing pretrained CNN models for stroke classification from CT images. The study employed well-established CNN architectures, including EfficientNet-B0, ResNet50, and VGG19, which were pretrained on large-scale datasets such as ImageNet dataset. These pretrained models were used to extract high-level features from brain CT scans. After that, these features were fed to traditional ML algorithms such as SVM and K-Nearest Neighbors (KNN) for classification into three classes (ischemic stroke, hemorrhagic stroke, and normal brain). By combining DL/TL-based feature extraction with ML classifiers, this hybrid approach achieved exceptional results, with the EfficientNet-B0 + SVM model achieving the highest accuracy of 95.13%, recall of 94.93%, precision of 95.06%, and an F1-score of 94.94%.

Another study (Yopiangga et al. 2024) utilized state-of-the-art vision transformers (ViT Base 16) for stroke classification. The study employed the pretrained VIT model to classify CT scan images into three categories:Not stroke, Hemorrhagic stroke, and Ischemic stroke. .The ViT model, which leverages the power of self-attention mechanisms, was fine-tuned to adapt to the stroke classification task, demonstrating its effectiveness. The experiment showed an accuracy score of 91% for the best classification model developed in this study. Although this study highlights the advantages of transformer-based pretrained models for medical image classification, mainly for tasks of stroke classification using CT scans, some advanced transformer-based pretrained models with effective capability to capture both local and global features have not been explored in addressing such a problem.

In the same direction, (Abbaoui et al. 2024) explored transfer learning with a Vision Transformer (ViT-B16) model for acute ischemic stroke identification from MRI scans. They fine-tuned a ViT-B16 (pretrained on ImageNet) on a 342 T1-weighted MRI images (stroke vs. normal cases from Moroccan hospitals) dataset. The MRI data underwent standardization and image augmentation (random flips and zoom transformations) to improve generalization. The ViT model's transformer encoder was combined with new dense layers and trained for 50 epochs, with optimization via Adam. This ViT-based model achieved an impressive accuracy of ~97.6% in classifying ischemic stroke, substantially higher than a conventional VGG16 CNN baseline (~90% accuracy) on the same MRI set. ViT-B16 outperformed multiple CNN benchmarks (e.g. ResNet50 at 87%, InceptionV3 at 82%) from a prior study. The model also demonstrated high precision and recall (≈96–98% for both classes). These results illustrate that transformer-based transfer learning can improve stroke detection accuracy over traditional CNN methods, underlining the potential of advanced pretrained models in medical imaging.

Additionally, a very recent study (Soni 2025) focused on hemorrhagic stroke (intracranial hemorrhage) detection in CT scans using transfer learning. The study leveraged the large RSNA intracranial hemorrhage CT dataset, fine-tuning a pretrained VGG-16 model (after considering other CNN backbones like AlexNet, EfficientNet-B2, ResNet50, MobileNet, and Inception) to classify hemorrhages into five subtypes plus normal. The VGG-16's final classification layer was modified to output six classes, and data augmentation techniques (e.g., rotations and contrast enhancement) were applied to improve robustness during fine-tuning. Despite the limited training data per class, the model achieved high performance – about 86.5% overall accuracy, with ~85.9% precision, ~86.2% recall, and ~86.7% F1-score on hemorrhage classification. This demonstrates robust transfer learning performance for hemorrhage detection on CT. The author noted some generalizability challenges (due to scanner and protocol variability) and plans to incorporate further enhancements, such as improved contrast

Despite the success that has been made toward stroke detection and classification, there are still some limitations or gaps that can be addressed for future studies. For example, although DL models have been developed to detect and classify strokes, the prediction performance is still limited. Furthermore, current models often fail to be generalized well when dealing with new and unseen datasets due to changes in the quality of CT images and different imaging protocols. Another limitation, most of the existing studies use CNNs to detect strokes using CT scans, with a few efforts to explore transformer-based models. Also, the lack of use of data enhancement techniques where advanced data augmentation techniques are still underused in stroke detection, which can overcome the issue of limited medical dataset size and diversity. The last gap to highlight is the Insufficient focus on the results interpretability of advanced AI models.

## 4 Methods

### 4.1 The proposed System Workflow

In this study, stroke identification is represented as supervised learning of multiclass classification, where the datasets consist of CT-scan images and the goal is to predict if the image is normal, ischemic stroke, or hemorrhagic stroke. The framework of the methodology is illustrated in Figure 1 and comprises seven main steps as follows:

1. **Data Acquisition:** to obtain the CT scans of the three classes.
2. **Data Preprocessing and augmentation:** The data is preprocessed to enhance image quality, and then data augmentation is applied using classical and GAN techniques.
3. **Transfer Learning of Pretrained Models:** the processed CT scan images are then used as inputs for the DL pretrained models to extract discriminative features automatically. These pretrained models are mainly ViT, MaxViT, ConvNext, and TNT.
4. **Classification:** Those auto-extracted features are fed into classification blocks using fully connected layers and a softmax function to produce an output in one of the defined classes (No Stroke, Hemorrhagic Stroke, or Ischemic Stroke).
5. **Explainable AI:** XAI clarifies the classification output, ensuring that the results are clear and convincing for physicians.

The subsequent sections explain each methodology component (e.g., framework step) in more detail.

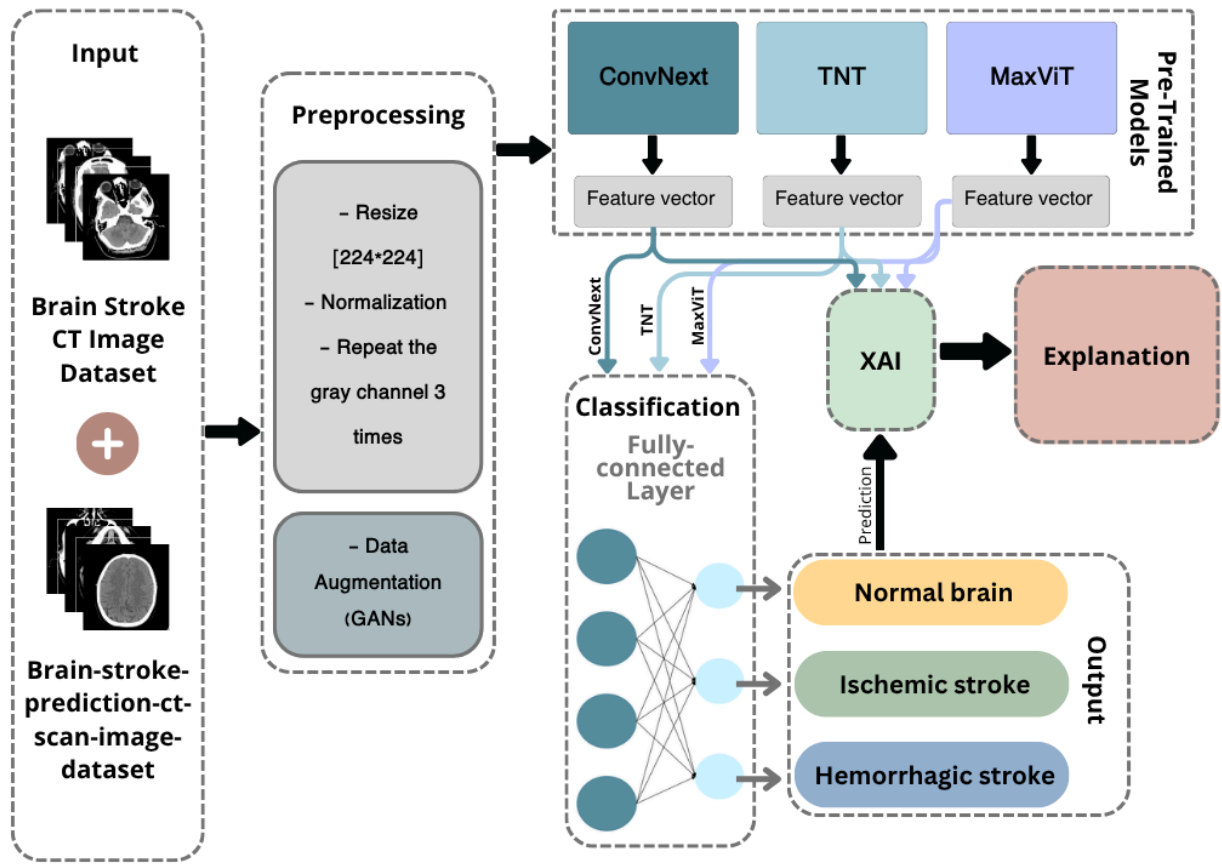

**Figure 1:** The framework of the proposed system of Stroke Classification.

## 4.3 Datasets

The stroke images utilized in this study are obtained from the dataset published in a recent study for brain stroke detection (Çınar, Kaya, and Kaya 2023). The original dataset was curated from data provided by the Republic of Turkey's Ministry of Health, specifically obtained through its e-Pulse and Teleradiology systems. The data, which includes brain CT images in both DICOM and PNG formats, was collected between 2019 and 2020. All images were anonymized to ensure patient privacy and comply with ethical data protection standards. Seven radiology specialists assigned this dataset through CT image descriptions, and the annotations were validated and confirmed by another radiologist to ensure the accuracy and clinical reliability.

In our study, we utilized the original PNG-format images available in the dataset, as they are well-suited for direct input into DL models without the need for DICOM decoding or conversion process. The dataset includes three classes, which are Ischemic stroke, Hemorrhagic stroke, and no stroke (normal brain scans). Table 2 provides a statistical summary of the dataset utilized in our experiments. As shown in the table, the classes of stroke are imbalanced, with the hemorrhagic and ischemic stroke categories having notably fewer samples compared to the normal class. Thus, to mitigate the impact of the imbalance issue, we applied advanced data augmentation techniques (i.e., GAN) on minor classes, "Hemorrhagic and Ischemic stroke," and used the weighted class function. The details of data augmentation are explained in the data preprocessing and augmentation system.

Table 2: Dataset statistics for brain stroke CT scan images

| Dataset Name | Type | Number of CT scans |
|---|---|---|
| **Republic of Turkey's Ministry of Health's** (Çınar, Kaya, and Kaya 2023). | Ischemic Stroke | 1,130 |
| | Hemorrhagic Stroke | 1,093 |
| | Normal (No Stroke) | 4,427 |
| | All classes | 6,650 |

### 4.4 Data Preprocessing

This study utilizes CT scan images with three DL pretrained models: TNT, MaxViT, and ConvNext. To ensure compatibility across these architectures, preprocessing and augmentation were carefully applied to standardize the input data format and optimize model performance. The standard image input size used in the pretrained models that rely on ImageNet data, such as ConvNeXt and TNT, is $224 \times 224$ pixels. Therefore, all images are resized to a fixed resolution [$224 \times 224$ pixels]. Additionally, since CT images are grayscale (single-channel) but pretrained models require three-channel (RGB) input, the grayscale channel was replicated across all three channels to meet the input specifications while avoiding color bias.

### 4.5 Data Augmentation

Data augmentation is a basic technique to address the issue of limited data in certain stroke categories, enhance model robustness, and reduce overfitting. In this research, we utilized a classical data augmentation and a conditional generative adversarial network (cGAN) techniques.

*4.5.1 Data Augmentation using Classical Techniques*

We applied several standard augmentation techniques during training to enrich the diversity of the data, improve generalization, and prevent overfitting. These included geometric transformations such as random cropping, horizontal flipping, and rotation, as well as photometric transformations like color jittering.

*4.5..2 Data Augmentation using conditional GAN (cGAN)*

Conditional GAN is an extension of the traditional GAN model, with conditional information added to both the generator and the discriminator, allowing the model to generate images based on class labels (Mirza and Osindero 2014). It generates CT scan images, specifically for minor classes such as hemorrhagic and ischemic stroke. These images are synthetic, but they mimic the original data.

The cGAN framework comprises two adversarial components: the generator and the discriminator. The generator generates fake data dependent on random noise with adaptive degeneration on the specific case or features. On the other hand, the discriminator evaluates the generated image by distinguishing between real and generated samples. Through the iterative training process, the generator progressively improves its output quality based on feedback from the discriminator. As a result, high-quality, class-specific synthetic images were produced and integrated into the training set to improve class balance and model generalization (Bok and Langr 2019).

**4.5 Pretrained Model Selection and Feature Extraction**

In our proposed stroke detection and classification method, we utilized a set of state-of-the-art pretrained models to efficiently extract discriminative features from CT scan images. These models, originally trained on large-scale datasets such as ImageNet (Krizhevsky, Sutskever, and Hinton 2017), have demonstrated an outstanding capability in capturing low-level patterns (e.g., edges and textures) and high-level semantic features (e.g., global context). We adapted these models to the medical imaging domain, specifically for classifying brain CT scans, in a transfer learning style. The selected models fall into two categories: First, Transformer-based: The Vision Transformer (ViT) and Transformer-in-Transformer (TNT). Second, Hybrid architectures that integrate convolutional and transformer mechanisms (e.g., ConvNeXt and MaxViT). Among these models,
- For benchmarking and comparative analysis: ViT serves as the baseline model for this study, so its performance can be used to evaluate the impact of more advanced architectures.
- TNT is included for its hierarchical dual-transformer structure, which models both intra-patch and inter-patch relationships,

- ConvNeXt represents a hybrid CNN-transformer design that integrates convolutional inductive biases with transformer-inspired components.
- Finally, MaxViT serves as the core model in our study due to its dual attention mechanisms and efficient feature representation capabilities. It stands out as a powerful architecture for medical image analysis, particularly for CT-based stroke detection, due to its ability to balance transformer level features with CNN-like efficiency and structure awareness.

### *4.5.1 Vision Transformer (ViT) Pretrained Model*

Inspired by the success of Transformers in NLP, the Vision Transformer (ViT) (Dosovitskiy et al. 2020) was developed specifically for computer vision tasks. ViT is the foundation of transformer-based architectures successfully applied to image classification tasks. This approach first divides an input image into multiple non-overlapping patches. Each patch is flattened and converted into a vector, followed by dimensionality reduction through a linear projection. A transformer encoder subsequently processes these vector embeddings as though they were token embeddings from an NLP model. This allows the ViT to capture local and global patterns in the image, which is crucial for analyzing CT scans where stroke-related features may be subtle and widely distributed. In this study, ViT is used as the baseline model to evaluate the impact of more advanced architectures (e.g., TNT and MaxViT) compared to the standard transformer approach.

### *4.5.2 Transformer in Transformer Pretrained Model*

The transformer architecture employed in this study includes the Transformer-in-Transformer (TNT) (Han et al. 2021) model, specifically the tnt_s_patch16_224 variant available through the Timm library. TNT, one of the most popular models, extends the traditional ViT by introducing a hierarchical attention mechanism that captures both fine-grained local features and high-level global context, which is particularly advantageous for analyzing complex medical images such as brain CT scans. TNT enhances the feature representation ability by dividing each input image into several patches as "visual sentences" and further subdividing each patch into sub-patches as "visual words". The TNT architecture consists of two core components: the first is the *inner transformer*, which models the relationship between visual words for local feature extraction (local representations). The second is the *outer transformer*, which captures the intrinsic information from the sequence of sentences (global representations), as shown in Figure 2. The TNT model was selected for this study because of its unique ability to extract local details and global relationships (Han et al. 2021). Each component of the TNT model plays a crucial role in transforming raw CT scan images into meaningful feature representations for classifying stroke types.

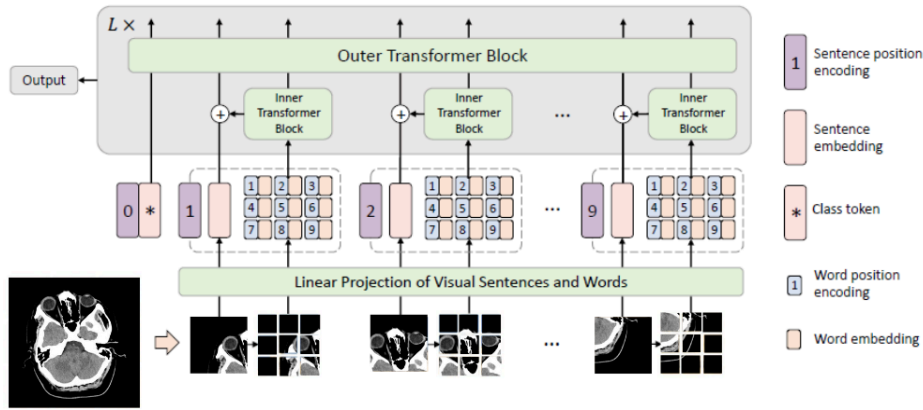

**Figure 2**: Transformer-iN-Transformer (TNT) architecture (Han et al. 2021).

*4.5.3 Convolutional Neural Network Next (ConvNeXt) Pretrained Model*

ConvNeXt is a modern convolutional architecture (Liu et al., n.d.), designed to incorporate several Transformer-inspired enhancements while preserving the efficiency and inductive biases of standard CNNs. In this study, we adopted the **convnext_base** variant from the timm library due to its high performance in various medical imaging tasks. These improvements of classic CNN include replacing the batch normalization with layer normalization, and increasing the convolutional kernel size (from 3x3 to 7x7) to approximate the global self-attention mechanism in transformers. These replacements enable the capture of more detailed information that are essential features for medical imaging tasks (Liu et al., n.d.). We selected ConvNeXt for stroke detection for the following key reasons:
- ConvNext performs exceptionally well in classifying medical imaging (e.g., histopathological and radiological images), making it highly suitable for analyzing brain CT scans.
- The ability to detect stroke-related features with the larger convolutional kernels (7x7) enables extracting discriminative and useful features from CT scans, including:
    - Blood Accumulation for Hemorrhagic Stroke: ConvNext is trained to identify high-intensity regions in CT scans, which indicate blood accumulation, a key characteristic of hemorrhagic strokes.

- Infarction Tissue for Ischemic Stroke: For ischemic strokes, ConvNext detects low-intensity regions that signify tissue infarction caused by inadequate blood flow.

### 4.5.4 MaxViT Hybrid Pretrained Model

The final model adopted in this study is the **Multi-Axis Vision Transformer (MaxViT) model**, proposed by (Tu et al. 2022) and implemented using the **timm** library. MaxViT is a hybrid vision transformer that combines local and global attention within the transformer framework while also incorporating CNN-like features into its architecture. This design allows MaxViT to benefit from both global and local receptive fields across the entire network. At the core of MaxViT is its **multi-axis attention design**, which incorporates two complementary attention types.

- **Grid Attention:** Operates within small patches of the image, focusing on capturing fine-grained local features essential for identifying subtle stroke-related abnormalities such as infarcts or localized hemorrhage.
- **Axial Attention:** Captures broader interactions along the horizontal and vertical axes, providing a global contextual understanding across the entire CT image.

MaxViT follows a hierarchical CNN-like design while embedding transformer-based attention mechanisms across multiple stages. We selected the MaxViT model for several reasons:

1. It balances extracting subtle local features with understanding the global context. In the case of stroke detection, the features are categorized as, local features that help detect subtle changes in brain structures that indicate strokes, and global context that ensures consideration of the broader anatomical structure, which is crucial for precisely classifying stroke types.
2. Built-in segmentation capability: in stroke detection, we often need to segment specific areas, such as brain lesions or blood clots, to classify the stroke type. However, due to its multi-axis attention, MaxViT can process both local and global features within the entire image, allowing it to analyze CT scans without needing a separate segmentation step, such as UNet. The MaxViT model inherently performs segmentation as part of its processing pipeline. Therefore, the segmentation step can be bypassed, as the model effectively handles it during the automatic feature extraction and classification process.

**4.6 Transfer Learning and Fine-Tuning for the Pretrained Models**

This research employed transfer learning (TL) to utilize the state-of-the-art pretrained models introduced previously in section 3.4. These pretrained models, originally trained on large-scale datasets like the ImageNet dataset (Krizhevsky, Sutskever, and Hinton 2017; Deng et al. 2009), serve as feature extractors by providing robust and generalizable representations of CT scan images, which is capable of capturing both low- and high-level visual representations. By applying TL, the pretrained models are fine-tuned to adapt their learned features to the domain-specific task, which is stroke detection and classification. This process ensures vital feature representation and significantly reduces the computational cost and training time compared to training models from scratch. It also improved the prediction accuracy for this application. To adapt the models to our clinical task, we fine-tuned each architecture (MaxViT, ConvNeXt, and TNT) through the following steps:

1. **Customizing the Final Layers:** The final layers of each model were adjusted to match stroke classification tasks. A new fully connected layer will replace the original output layers, adapted for our task. The classifier contains **three output neurons**, each corresponding to one of the target categories: **Class 0:** No Stroke, **Class 1:** Hemorrhagic Stroke, or **Class 2:** Ischemic Stroke.
2. **Freezing and Fine-Tuning the Layers:** To balance computational efficiency with performance. For each model, all layers were frozen except for the final classification layer, in order to take advantage of the pretrained weights in the feature extraction stages, while allowing the classification layer to adapt to the characteristics of the new data.
3. **Hyperparameter Configuration:** For hyperparameter optimization, The models were trained with a variety of hyperparameter settings (see Table 3). The most important parameters we have optimized include batch size, learning rate, the optimizer, and the number of epochs. We have configured the hyperparameters for all models. These configurations were experimentally tuned to optimize each model's performance on the CT stroke classification task.

**Table 3:** Hyperparameter Configuration. Bold indicates the selected values.

| Hyperparameter | Values |
|---|---|
| Optimizer | **Adam**, mAdam |
| Learning Rate | 1e-3, 1e-5, **3e-4** |
| Batch Size | 16, 32, **64** |
| Number of Epochs | **25**, 40, 50, 100 |
| Drop out ratio | 0.03, 0.04, **0.05** |

## 4.7 Classification

Since our problem is a multiclass classification, we utilized a fully connected layer with three output neurons to be added to each pretrained model. Each neuron corresponds to one of the three target classes: No Stroke, Hemorrhagic Stroke, and Ischemic Stroke. The SoftMax function, the most commonly used activation function for multiclass classification, was employed to compute the probability distribution across the three classes, enabling the model to output the likelihood of each class for a given input data sample (i,e., CT scan image).

## 4.8 Explainable AI (XAI) Integration

We utilized XAI techniques aimed at making the model's predictions more interpretable, reliable, and transparent. Grad-CAM is a technique used in XAI. It is used to generate visual explanations for CNNs by highlighting the most important regions in the input image. The important regions have contributed significantly to model predictions. Grad-CAM depends on calculating the gradients of the result of a class with respect to the feature maps of a convolutional layer to produce a heatmap highlighting the most important regions (Selvaraju et al. 2016). This approach is useful when we use an AI model and need to explain the results, especially in the medical field. In medical images, using Grad-CAM is helpful. It allows doctors to understand the decision-making process of DL, and it helps use AI systems in healthcare applications (Singh, Sengupta, and Lakshminarayanan 2020). In this research, we use CT images for the early detection and classification of strokes. Grad-CAM can explain the model predictions to physicians as it can identify the areas of a CT image that affected the classification, such as infarcted or hemorrhagic areas.

# 5 Experiments and Evaluation Protocols

We outline the experimental design used to evaluate the performance of the proposed AI-based framework for stroke detection and classification. Also, we describe the setup of the training and testing process, including the dataset split, training protocols, and key evaluation metrics.

## 5.1 Evaluation Metrics

In this study, the problem is a multiclass classification task to predict the type of stroke *(No Stroke, Hemorrhagic Stroke, or Ischemic Stroke)*. Evaluation metrics are an integral part of the model as they assess its performance and help verify its efficiency. To evaluate our models, we utilized Accuracy, Precision, Recall, and F1-score metrics. These metrics are derived from the confusion matrix, which consists of four key components: True Positive (**TP**), True Negative (**TN**), False Positive (**FP**), and False Negative (**FN**), as shown in *Figure 9*.

**Accuracy**: One of the most important metrics used in classification tasks. It represents the ratio of correctly predicted instances (TP and TN) to the total number of instances. TP and TN are correctly predicted, while FP and FN are incorrectly predicted. The formula for accuracy is as follows:

$$Accuracy = \frac{TP + TN}{TP + TN + FP + FN}$$

**Precision**: This metric indicates how many predicted positive values are actually positive. It is the ratio of TP to (TP + FP). The formula for precision is as follows:

$$Precision = \frac{TP}{TP + FP}$$

**Recall**: Recall measures how well the model identifies true positive instances (e.g., detecting all actual stroke cases). It is the ratio of TP to (TP + FN). The formula for recall is as follows:

$$Recall = \frac{TP}{TP + FN}$$

**F1 Score**: The F1-score depends on both recall and precision metrics. It is typically a harmonic mean of precision and recall, used when a balance between the two is essential. The formula for the F1 Score is as follows:

$$F1\ Score = 2 * \left(\frac{Precision * Recall}{Precision + Recall}\right)$$

## 5.2 Data Splitting Protocols

To assess the performance of our proposed models and ensure consistency with the baseline methods, a standard train-test splitting strategy was utilized in this study. This approach facilitates a direct and fair comparison by following the same splitting procedure as the baseline studies. Moreover, to maintain the original class distribution and address potential class imbalance, the splitting was applied in a stratified style, where each class *(No Stroke, Hemorrhagic Stroke, and Ischemic Stroke)* is proportionally represented in both the training and testing sets. The data was split as follows:

- **Training set:** 80% of the data was used to train the models to classify the data into the three categories.
- **Testing set:** 20% of the data was reserved to evaluate the performance of the final model on unseen test data.

## 5.3 Data Augmentation using cGAN Experiment

To address the class imbalance in the dataset, where the number of images in the Hemorrhagic and Ischemic stroke categories was significantly lower than that in the normal category, we employed cGAN to generate synthetic images that enhance the representation of the underrepresented classes. The generator was designed to take a random noise vector with the class label (Normal, Ischemic, or Hemorrhagic) as input and then generate a grayscale image based on the class. The generator consists of several Dense and Conv2DTranspose layers aimed at realistically reconstructing the visual structure of the image. The discriminator received both real and generated images and their associated class labels, distinguishing between real and fake samples. It consists of a set of Conv2D layers followed by Dense layers to classify the input as real or fake. During training, the generator and the discriminator are alternately updated using their respective loss functions. The generator learns to produce images capable of deceiving the discriminator, while the discriminator is trained to enhance its ability to detect fake images. The model was trained in two distinct phases:

1. **Phase 1 (Stabilization Phase)**: The cGAN was trained for 200 epochs, and the quality of the generated images stabilized noticeably. However, these images were not saved during this phase.
2. **Phase 2 (Generation Phase):** With the image-saving mechanism activated, the training continued for an additional 800 epochs. At each epoch, approximately 800 synthetic images were generated to balance the dataset (split evenly between the Ischemic and Hemorrhagic categories). These images were automatically categorized and stored in class-specific folders.

After the completion of training, the quality of the generated images was assessed, showing a significant similarity to the original CT scans. Based on this assessment, synthetic images for the Hemorrhagic and Ischemic categories generated from Phase 2 were combined with the original training data for these two categories. In contrast, no synthetic images for the Normal category were merged to maintain a balance in the distribution among categories. Figure 3 illustrates a sequence of synthetic CT images generated using the cGAN framework for both Ischemic (top row) and Hemorrhagic (bottom row) stroke classes. Using the cGAN model. The images are captured at various training epochs, allowing visual examination of the model's progression in learning crucial stroke features. The synthetic images from earlier epochs (such as epochs 125 and 130) show blurred anatomical structures, which suggest early-stage fluctuation and mode failure problems. Image quality significantly improves at later epochs in training (after 500 epochs for ischemic and after 664 epochs for hemorrhagic).

Overall, the qualitative advancement demonstrates that cGAN effectively captures modality-specific stroke features, contributing to improved class balance and enhanced multiclass stroke classification.

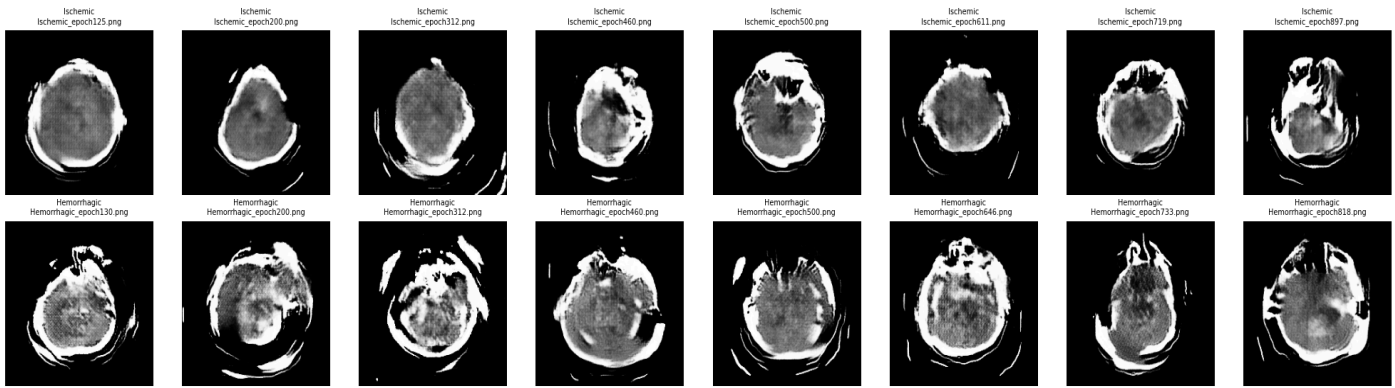

**Figure 3:** Examples of Synthetic CT Scan Images Generated by the cGAN for Hemorrhagic and Ischemic Classes

## 5.4 Implementation Details and Resources

We implemented our proposed method using a combination of local-based and cloud-based computational resources. Local development and preliminary testing were conducted on an ASUS X415JF laptop equipped with an Intel Core i7 (10th generation) processor, NVIDIA GeForce MX330 GPU, 16 GB RAM, and a 1 TB SSD. For full-scale training and fine-tuning, we leveraged Google Colab Pro, which offers access to high-performance GPUs (such as Tesla T4 and P100) and extended RAM. This environment was chosen for its accessibility and scalability, significantly reducing training time and improving model efficiency. All pretrained models (e.g., TNT, MaxViT, ConvNeXt) were loaded from the timm library (PyTorch Image Models), a reliable and popular resource that offers access to hundreds of cutting-edge vision models pre-trained on ImageNet. This library supports seamless integration and transfer learning functionalities, which are essential for our work. The Torchvision library was used for data processing and transformations, and scikit-learn for calculating performance metrics such as Precision, Recall, and F1-score.

## 6 Results and Discussion

### 6.1 AI Model Prediction Performance

We present and discuss the prediction performance results obtained from four transformer-based and hybrid pretrained models used in this study. Each model was trained using 80% of the dataset and evaluated on the remaining 20% of unseen test data using standard classification metrics, including accuracy, precision, recall, F1-score, and loss, as summarized in Table 6. The primary objective is to classify CT scans into one of three categories: hemorrhagic stroke, ischemic stroke, or normal (no stroke). To evaluate the impact of data augmentation on model performance, we conducted three experimental settings for each model: training on the original dataset, training with classical data augmentation, and training with synthetic images generated using a conditional GAN (cGAN).

As shown in Table 3, all models demonstrated strong predictive performance across the three experiments. MaxViT consistently achieved the highest results, while ViT recorded the lowest across all metrics. This may be because ViT's relatively simple architecture lacks the hierarchical and multi-scale attention mechanisms in more advanced models like MaxViT and TNT. These features allow the latter models to extract finer details from CT scans, essential for accurate stroke classification.

The baseline (non-augmented) versions of TNT, ConvNeXt, and MaxViT achieved competitive accuracies of 93.83%, 95.64%, and 97.67%, respectively. However, each model showed consistent improvements when trained with classical data augmentation, and further gains were observed with cGAN-based augmentation. This progression highlights the clear benefit of using data augmentation, especially CGAN generated synthetic CT scans, to improve model generalization and address class imbalance. The effect of cGAN augmentation was most prominent for TNT and ConvNeXt, which improved their accuracy by approximately 2% compared to their baseline versions. In contrast, MaxViT showed only a low improvement, increasing to 98.00%. This limited gain is likely due to MaxViT's robust architecture, which combines block attention, grid attention, and convolutional layers to learn efficiently even from smaller datasets.

Table 3: Prediction Performance of Pretrained Models on the Multiclass Stroke Classification.

**Bold** font indicates the best-performing models and *Italic* indicates the second best performing model.

| DL Models | Accuracy | Loss Value | F1-score | recall | precision |
|---|---|---|---|---|---|
| ViT | 0.8662 | 1.5910 | 0.86 | 0.87 | 0.86 |
| ViT + classical DA | 0.8556 | 2.6897 | 0.85 | 0.86 | 0.85 |
| **ViT + cGAN** | 0.8964 | 0.9478 | 0.90 | 0.90 | 0.90 |
| TNT | 0.9286 | 0.6686 | 0.93 | 0.93 | 0.93 |
| TNT + classical DA | 0.9383 | 0.5142 | 0.94 | 0.94 | 0.94 |
| **TNT + cGAN** | 0.9600 | 0.2830 | 0.96 | 0.96 | 0.96 |
| ConvNext | 0.9617 | 0.2821 | 0.96 | 0.96 | 0.96 |
| ConvNext + classical DA | 0.9564 | 0.3558 | 0.96 | 0.96 | 0.96 |
| **ConvNext + cGAN** | 0.9745 | 0.1732 | 0.97 | 0.97 | 0.97 |
| MaxViT | 0.9759 | 0.2132 | 0.98 | 0.98 | 0.98 |
| MaxViT + classical DA | *0.9789* | *0.1339* | *0.98* | *0.98* | *0.98* |
| **MaxViT + cGAN** | **0.9800** | **0.1073** | **0.98** | **0.98** | **0.98** |

In summary, the best results were achieved using the cGAN-augmented datasets: MaxViT achieved the highest accuracy of 98.00% and other metrics, followed by ConvNeXt with 97.45%, and TNT with 96.00%. These findings prove that transformer-based models benefit from data augmentation strategies, particularly those with more advanced architectures.

## 6.2 Comparison with the state-of-the-art methods

To investigate our method's prediction performance and illustrate its effectiveness, we aim to compare it with a study titled *Classification of brain ischemia and hemorrhagic stroke using a hybrid method (Çınar, Kaya, and Kaya 2023)*, which is, as far as we know, the only work focused on predicting multiclass stroke prediction, making it the baseline method of our study. However, this method utilized different datasets we plan to obtain for a fair comparison. This study used a pretrained model, EfficientNet-B0, with an SVM classifier and achieved. The data were obtained from the Turkish Ministry of Health's "e-Nabz" and "Remote Radiology" systems, where the images were classified by specialized radiologists into three categories: ischemic stroke, hemorrhagic stroke, and normal cases. The data preprocessing included steps such as resizing, contrast enhancement, noise removal, and normalization, aimed at preparing the images for effective processing by neural networks. The hybrid model in the study relies on utilizing DL capabilities to extract representative features from the images, and then passing these features to an SVM model to complete the classification process in a flexible and high-generalization manner. In our study, to conduct a fair comparison, we utilized the same dataset, problem formulation, data splitting protocol, and evaluation metrics across all models. Table 4 shows the comparative results, which demonstrate that our proposed method utilizing three advanced transformer-based and hybrid-based models consistently outperforms the state-of-the-art method.

**Table 4:** Comparative Performance Between the Proposed Transformer-Based Models and the Baseline EfficientNet-B0+SVM Mode. *Bold font indicates the best-performing model.*

| Study | Models | **Accuracy** | **F1-score** | **recall** | **Precision** |
|---|---|---|---|---|---|
| (Çınar, Kaya, and Kaya 2023) | EfficientNet-B0 + SVM | 95.13% | 94.94% | 94.93% | 95.06% |
| Our proposed method in this study | *MaxViT + cGAN* | **98.00%** | **98.00%** | **98.00%** | **98.00%** |
| | ConvNext + cGAN | 97.45% | 97.00% | 97.00% | 97.00% |
| | TNT + cGAN | 96.00% | 96.00% | 96.00% | 96.00% |

## 6.3 Explainable AI Results and Insights

In sensitive domains such as the medical field, the integration of AI models must go beyond accurate classification results and quantitative metrics. It must also understand the reasons behind decision-making and give transparency and interpretability. XAI helps clarify the features or areas within CT-scan images that the model relied upon during classification. In this study, we applied XAI to the best-performing model, MaxViT, using the Grad-CAM++ technique. This well-established method highlights the important areas within the convolutional layers contributing to model prediction. We examined three different layers within the model to evaluate their ability to focus on areas of importance:

- Early Layer: **stem.conv1**
- Mid-Level Feature Layer: **stages.1.blocks.1.conv.conv2_kxk**
- Deep Layer: **stages.3.blocks.1.conv.conv2_kxk**

**Figure 4 (a) and (b)** shows Grad-CAM++ visualizations across these layers for two representative CT scan samples (one hemorrhagic (top row) and one ischemic (bottom row). The first column shows the original input image. The second column displays the early layer's activation that highlights broad and low-relevance regions, which does not show any actual focus on the important brain areas. The third column presents a mid-level layer with partially relevant activation but scattered focus. The fourth column corresponds to the deep layer, which accurately focuses on the core brain lesion areas related to correct classification, making it the most suitable for interpreting the model's decisions. This visualization reveals that deeper layers provide more clinically meaningful and focused explanations.

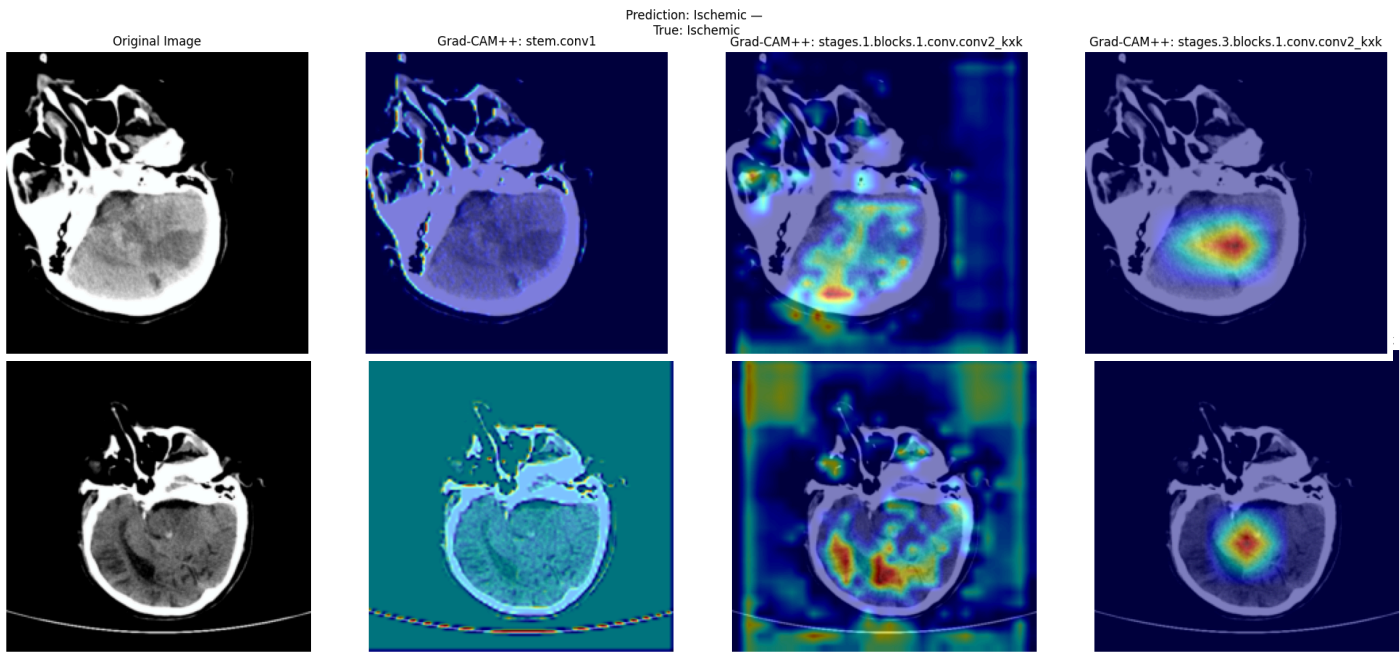

**Figure 4: (a)** Grad-CAM++ visualizations across three convolutional layers on CT scan samples from the Hemorrhagic class. **(b)** Grad-CAM++ visualizations across three convolutional layers on CT scan samples from the Ischemic class.

Additionally, as illustrated in **Figure 5**, results showed that the model tended to focus more accurately on critical areas in Hemorrhagic cases than in Ischemic cases, indicating differences in the model's response to each type of image.

To better understand the heatmaps on the CT scans in both Figures 4 and 5, besides the **Layer-wise comparison** above, we analyzed them through the following key aspects:

- **Color intensity:** The color gradient, ranging from blue to red, indicates the degree to which a region contributed to the model's prediction. The red color in heatmaps indicates the severity of importance in the model's decision, as the darker the red color in a specific area, the more the model relies on that area in making its final decision. In contrast, the **blue areas** indicate little to no contribution.
- **Shape and Localization:** Heatmaps with focused dark red regions suggest confident and localized attention, likely corresponding to stroke-affected areas.
- **Alignment with Stoke Lesion Regions:** Although no separate segmentation mask technique was used, visual inspection indicates that the darkest red areas in the heatmaps often align with radiological signs of hemorrhage or ischemia. This overlap supports the validity of the model's focus and adds confidence in its interpretability.

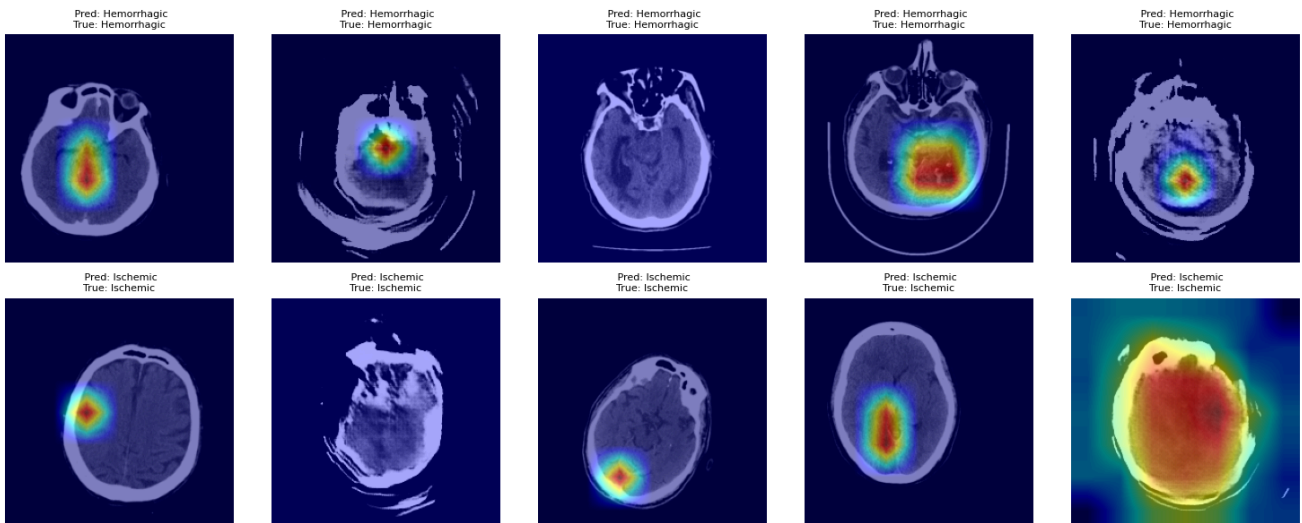

**Figure 5:** Comparison of Grad-CAM++ heatmaps for multiple samples from the Hemorrhagic and Ischemic classes.

# 7 Conclusion

This research focused on the early detection and classification of stroke types (ischemic, hemorrhagic, or no stroke) using CT scan images. The significance of this work lies in its contribution to developing early stroke detection systems, which are crucial due to the severe health, social, and economic consequences of strokes. Additionally, this study emphasized using CT scans, which are widely available in medical facilities. Throughout this research, we reviewed previous studies on stroke detection and analyzed literature on different models to identify limitations, advantages, and gaps in existing DL techniques. The problem was framed as a multiclass classification task to achieve the research objectives, and the dataset was preprocessed to align with the proposed models.

      We implemented and evaluated several transformer-based and Hybrid-based architectures, such as TNT, MaxTiV, and ConvNext, to assess their effectiveness in stroke classification. We utilized ViT as the baseline model to benchmark performance. This consistent focus on transformer-based models allowed us to demonstrate their ability and suitability for stroke detection tasks using CT imaging. Experimental results in terms of multiple evaluation metrics confirmed that incorporating transformer-based components into DL models (MaxViT, TNT, and ConvNext) achieved high predictive performance, with significant improvements when combined with data augmentation techniques. For future development to enhance the scope and impact of this research, we recommend the following direction:

- Expanding data sources by incorporating larger and more diverse datasets.
- Advancing towards real-time stroke detection systems in clinical environments.
- Implement attention-based XAI techniques to enhance the interpretability of model decisions further, leveraging the inherent attention mechanisms in transformer-based architectures to provide clearer insights.
- Integrate additional data, such as vital signs, with CT scans to improve model performance.

This research provides valuable insights into improving long-term health conditions, reducing mortality rates, and supporting healthcare professionals in managing brain stroke.

## Data Availability

The datasets used for this research are available upon request.

## Competing interests

The authors have declared that no conflict of interests exists.

## References:

Abbaoui, Wafae, Sara Retal, Soumia Ziti, and Brahim El Bhiri. 2024. "Automated Ischemic Stroke Classification from MRI Scans: Using a Vision Transformer Approach." *Journal of Clinical Medicine* 13 (8): 2323.

Alamro, Hind, Maha A. Thafar, Somayah Albaradei, Takashi Gojobori, Magbubah Essack, and Xin Gao. 2023. "Exploiting Machine Learning Models to Identify Novel Alzheimer's Disease Biomarkers and Potential Targets" 13 (1): 4979.

Albaradei, Somayah, Abdurhman Albaradei, Asim Alsaedi, Mahmut Uludag, Maha A. Thafar, Takashi Gojobori, Magbubah Essack, and Xin Gao. 2022. "MetastaSite: Predicting Metastasis to Different Sites Using Deep Learning with Gene Expression Data." *Frontiers in Molecular Biosciences* 9 (July):913602.

Albaradei, Somayah, N. Alganmi, Abdulrahman Albaradie, Eaman Alharbi, Olaa Motwalli, Maha A. Thafar, Takashi Gojobori, M. Essack, and Xin Gao. 2023. "A Deep Learning Model Predicts the Presence of Diverse Cancer Types Using Circulating Tumor Cells." *Scientific Reports* 13 (1): 21114.

Albaradei, Somayah, Maha A. Thafar, Asim Alsaedi, Christophe Van Neste, Takashi Gojobori, Magbubah Essack, and Xin Gao. 2021. "Machine Learning and Deep Learning Methods That Use Omics Data for Metastasis Prediction," September. https://scholar.google.com/citations?view_op=view_citation&hl=en&citation_for_view=QgCWtUQ AAAAJ:Zph67rFs4hoC.

Alghoraibi, Huda, Nuha Alqurashi, Sarah Alotaibi, Renad Alkhudaydi, Bdoor Aldajani, Lubna Alqurashi, Jood Batweel, and Maha A. Thafar. 2025. "A Smart Healthcare System for Monkeypox Skin Lesion Detection and Tracking." *arXiv [cs.CV]*. arXiv. http://arxiv.org/abs/2505.19023.

Alharthi, Asrar G., and Salha M. Alzahrani. 2023. "Multi-Slice Generation sMRI and fMRI for Autism Spectrum Disorder Diagnosis Using 3D-CNN and Vision Transformers." *Brain Sciences* 13 (11). https://doi.org/10.3390/brainsci13111578.

Alzahrani, Salha M. 2025. "MS-Trust: A Transformer Model with Causal-Global Dual Attention for Enhanced MRI-Based Multiple Sclerosis and Myelitis Detection." *Complex & Intelligent Systems* 11 (8). https://doi.org/10.1007/s40747-025-01945-2.

Bok, Vladimir, and Jakub Langr. 2019. *GANs in Action: Deep Learning with Generative Adversarial Networks*. Simon and Schuster.

Çınar, Necip, Buket Kaya, and Mehmet Kaya. 2023. "Classification of Brain Ischemia and Hemorrhagic Stroke Using a Hybrid Method." In *2023 4th International Conference on Data Analytics for Business and Industry (ICDABI)*, 5:279–84. IEEE.

Deng, Jia, Wei Dong, Richard Socher, Li-Jia Li, Kai Li, and Li Fei-Fei. 2009. "ImageNet: A Large-Scale Hierarchical Image Database." In *2009 IEEE Conference on Computer Vision and Pattern Recognition*, 248–55. IEEE.

Denslow, Elizabeth. 2022. "Understanding Your Prognosis After Stroke: Factors That Affect Recovery." Flint Rehab. July 19, 2022. https://www.flintrehab.com/stroke-recovery-prognosis/.

Dosovitskiy, Alexey, Lucas Beyer, Alexander Kolesnikov, Dirk Weissenborn, Xiaohua Zhai, Thomas Unterthiner, Mostafa Dehghani, et al. 2020. "An Image Is Worth 16x16 Words: Transformers for Image Recognition at Scale." *arXiv [cs.CV]*. arXiv. https://bibbase.org/service/mendeley/bfbbf840-4c42-3914-a463-19024f50b30c/file/264ac473-27b7-bd53-3963-f6a07df9b72e/Dosovitskiy_et_al___2021___An_Image_is_Worth_16x16_Words_Transformers_for_Im.pdf.pdf.

Flanders, Adam E., Luciano M. Prevedello, George Shih, Safwan S. Halabi, Jayashree Kalpathy-Cramer, Robyn Ball, John T. Mongan, et al. 2020. "Construction of a Machine Learning Dataset through Collaboration: The RSNA 2019 Brain CT Hemorrhage Challenge." *Radiology. Artificial Intelligence* 2 (3): e190211.

Gautam, Anjali, and Balasubramanian Raman. 2021. "Towards Effective Classification of Brain Hemorrhagic and Ischemic Stroke Using CNN." *Biomedical Signal Processing and Control* 63 (102178): 102178.

"Global Health Estimates: Life Expectancy and Leading Causes of Death and Disability." n.d. Accessed December 15, 2024. https://www.who.int/data/gho/data/themes/mortality-and-global-health-estimates.

Gomes, Joao, and Ari Marc Wachsman. 2013. "Types of Strokes." In *Handbook of Clinical Nutrition and Stroke*, 15–31. Totowa, NJ: Humana Press.

Han, Kai, An Xiao, E. Wu, Jianyuan Guo, Chunjing Xu, and Yunhe Wang. 2021. "Transformer in Transformer." Edited by M. Ranzato, A. Beygelzimer, Y. Dauphin, P. S. Liang, and J. Wortman Vaughan. *Neural Information Processing Systems* 34 (February):15908–19.

Karthik, R., R. Menaka, Annie Johnson, and Sundar Anand. 2020. "Neuroimaging and Deep Learning for Brain Stroke Detection - A Review of Recent Advancements and Future Prospects." *Computer Methods and Programs in Biomedicine* 197 (105728): 105728.

Kaya, Buket, and Muhammed Önal. 2023. "A CNN Transfer Learning‑based Approach for Segmentation and Classification of Brain Stroke from Noncontrast CT Images." *International Journal of Imaging Systems and Technology* 33 (4): 1335–52.

Krizhevsky, Alex, Ilya Sutskever, and Geoffrey E. Hinton. 2017. "ImageNet Classification with Deep Convolutional Neural Networks." *Communications of the ACM* 60 (6): 84–90.

Liu, Z., H. Mao, C. Y. Wu, and C. Feichtenhofer. n.d. "A Convnet for the 2020s." http://openaccess.thecvf.com/content/CVPR2022/html/Liu_A_ConvNet_for_the_2020s_CVPR_2022_paper.html.

Manzari, Omid Nejati, Hamid Ahmadabadi, Hossein Kashiani, Shahriar B. Shokouhi, and Ahmad Ayatollahi. 2023. "MedViT: A Robust Vision Transformer for Generalized Medical Image Classification." *arXiv [cs.CV]*. arXiv. http://arxiv.org/abs/2302.09462.

Mirza, Mehdi, and Simon Osindero. 2014. "Conditional Generative Adversarial Nets." *arXiv [cs.LG]*. arXiv. http://arxiv.org/abs/1411.1784.

Montaño, Arturo, Daniel F. Hanley, and J. Claude Hemphill 3rd. 2021. "Hemorrhagic Stroke." Edited by Steven W. Hetts and Daniel L. Cooke. *Handbook of Clinical Neurology* 176:229–48.

Peixoto, Solon Alves, and Pedro Pedrosa Rebouças Filho. 2018. "Neurologist-Level Classification of Stroke Using a Structural Co-Occurrence Matrix Based on the Frequency Domain." *Computers & Electrical Engineering: An International Journal* 71 (October):398–407.

Powers, William J. 2020. "Acute Ischemic Stroke." *The New England Journal of Medicine* 383 (3): 252–60.

Qiu, Wu, Hulin Kuang, Ericka Teleg, Johanna M. Ospel, Sung Il Sohn, Mohammed Almekhlafi, Mayank Goyal, Michael D. Hill, Andrew M. Demchuk, and Bijoy K. Menon. 2020. "Machine Learning for Detecting Early Infarction in Acute Stroke with Non-Contrast-Enhanced CT." *Radiology* 294 (3): 638–44.

Selvaraju, Ramprasaath R., Michael Cogswell, Abhishek Das, Ramakrishna Vedantam, Devi Parikh, and Dhruv Batra. 2016. "Grad-CAM: Visual Explanations from Deep Networks via Gradient-Based Localization." *arXiv [cs.CV]*. arXiv. http://arxiv.org/abs/1610.02391.

Shafaat, Omid, and Houman Sotoudeh. 2023. "Stroke Imaging." In *StatPearls [Internet]*. StatPearls Publishing.

Singh, Amitojdeep, Sourya Sengupta, and Vasudevan Lakshminarayanan. 2020. "Explainable Deep Learning Models in Medical Image Analysis." *arXiv [cs.CV]*. arXiv. http://arxiv.org/abs/2005.13799.

Sirsat, Manisha Sanjay, Eduardo Fermé, and Joana Câmara. 2020. "Machine Learning for Brain Stroke: A Review." *Journal of Stroke and Cerebrovascular Diseases: The Official Journal of National Stroke

*Association* 29 (10): 105162.

Soni, J. 2025. "Toward the Detection of Intracranial Hemorrhage: A Transfer Learning Approach." *Artificial Intelligence Surgery* 5 (2): 221–38.

Thafar, Maha A., Somayah Albaradei, Mahmut Uludag, Mona Alshahrani, Takashi Gojobori, Magbubah Essack, and Xin Gao. 2023. "OncoRTT: Predicting Novel Oncology-Related Therapeutic Targets Using BERT Embeddings and Omics Features." *Frontiers in Genetics* 14 (April):1139626.

Tu, Zhengzhong, Hossein Talebi, Han Zhang, Feng Yang, Peyman Milanfar, Alan Bovik, and Yinxiao Li. 2022. "MaxViT: Multi-Axis Vision Transformer." *arXiv [cs.CV]*. arXiv. https://doi.org/10.1007/978-3-031-20053-3_27.

Wang, Xiyue, Tao Shen, Sen Yang, Jun Lan, Yanming Xu, Minghui Wang, Jing Zhang, and Xiao Han. 2021. "A Deep Learning Algorithm for Automatic Detection and Classification of Acute Intracranial Hemorrhages in Head CT Scans." *NeuroImage. Clinical* 32 (102785): 102785.

Yopiangga, Alfian Prisma, Tessy Badriyah, Iwan Syarif, and Nur Sakinah. 2024. "Stroke Disease Classification Using CT Scan Image with Vision Transformer Method." In *2024 International Electronics Symposium (IES)*, 8:436–41. IEEE.